\documentclass{article} 
\usepackage{iclr2025_conference,times}

\usepackage{amsmath,amsfonts,bm}

\def\eqref#1{equation~\ref{#1}}

\def\1{\bm{1}}

\DeclareMathAlphabet{\mathsfit}{\encodingdefault}{\sfdefault}{m}{sl}
\SetMathAlphabet{\mathsfit}{bold}{\encodingdefault}{\sfdefault}{bx}{n}

\newcommand{\E}{\mathbb{E}}

\newcommand{\R}{\mathbb{R}}
\newcommand{\N}{\mathbb{N}}

\usepackage{hyperref}
\usepackage{url}
\usepackage{graphicx}
\usepackage{setspace}
\usepackage{float}
\usepackage{subcaption}

\newcommand{\repourl}{\url{https://github.com/lisa-wm/shortcuts_uncertainty.git}}

\title{Trust Me, I Know the Way: Predictive Uncertainty in the Presence of Shortcut Learning}

\author{Lisa Wimmer, Bernd Bischl \& Ludwig Bothmann \\
Department of Statistics, LMU Munich; 
Munich Center for Machine Learning (MCML) \\
\texttt{\{firstname.lastname\}@stat.uni-muenchen.de} \\
}

\newcommand{\lisa}[1]{\textcolor{teal}{TODO@LW: #1}}

\iclrfinalcopy 
\begin{document}

\maketitle

\begin{abstract}
The correct way to quantify predictive uncertainty in neural networks remains a topic of active discussion.
In particular, it is unclear whether the state-of-the art entropy decomposition leads to a meaningful representation of model, or \textit{epistemic}, uncertainty (EU) in the light of a debate that pits \textit{ignorance} against \textit{disagreement} perspectives.
We aim to reconcile the conflicting viewpoints by arguing that both are valid but arise from different learning situations.
Notably, we show that the presence of \textit{shortcuts} is decisive for EU manifesting as disagreement.
\end{abstract}

\begin{figure}[h]
    \centering \includegraphics[width=0.7\textwidth]{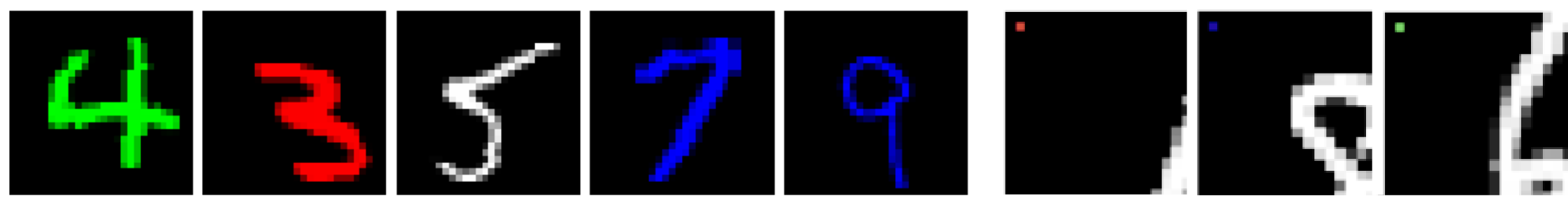}
    \caption{Examples of 3-class CMNIST3 data (\textit{left}) with full-image color shortcut and PMNIST3 (\textit{right}; cropped for better visibility) with $1 \times 1$ colored patches (shortcuts in 95\% of images).}
    \label{fig:example-data}
\end{figure}

\section{Introduction}
\label{sec:intro}

Safety-critical applications require models to report their uncertainty about predicted outcomes. 
The rise of deep learning, with its powerful-but-overconfident algorithms, has sparked rapid progress in uncertainty quantification  \citep[UQ,][]{guo_2017_CalibrationModern, gawlikowski_2023_SurveyUncertainty}.
Yet, the problem of faithfully representing, measuring and communicating uncertainty remains far from solved \citep{vanderbles_2019_CommunicatingUncertainty, bickfordsmith_2024_RethinkingAleatoric}.
One disputed aspect is the correct approach to disaggregate uncertainty into its \textit{aleatoric (AU)} and \textit{epistemic (EU)} components \citep{hullermeier_2021_AleatoricEpistemic, gruber_2023_SourcesUncertainty}.
AU arises when the features $X$ do not suffice to predict the target beyond doubt---e.g., due to omitted variables---even if we had access to infinite data.
The ground-truth distribution over labels $Y$, $\bm{\theta}^\ast = p(Y | X = \bm{x})$, will thus generally have positive dispersion (i.e., AU).
EU, on the other hand, surrounds the approximation quality of the model $h(\bm{x}) = \hat p_n(Y | X = \bm{x})$ based on $n$ observations.
EU is reducible in the sense that $\hat p_\infty$ recovers $p$\footnote{
We often assume a correctly specified model, equating hypotheses with parameterizations \citep{draper_1995_AssessmentPropagation}. 
}.

Attributing predictive uncertainty to its sources helps understand learning dynamics and provides an inroad for targeted model improvement. 
The widely-used decomposition of Shannon entropy\footnote{
Entropy is typically used with categorical target distributions.
An analogous decomposition exists for variance-based UQ in regression tasks \citep{depeweg_2018_DecompositionUncertainty, sale_2024_LabelwiseAleatoric}.
} \citep{shannon_1948_MathematicalTheory} as a measure of total uncertainty  (TU) offers a neat mathematical expression but has attracted criticism for conflating distinct concepts \citep{wimmer_2023_QuantifyingAleatoric, schweighofer_2024_InformationTheoreticMeasures, bickfordsmith_2024_RethinkingAleatoric} and being ineffective in practice \citep{mucsanyi2024benchmarkinguncertainty, fellaji_2024_EpistemicUncertainty}.
It is unclear, in particular, how well the resulting EU measure reflects lack of knowledge.
Besides the uncertainty of a single model $h \in \mathcal{H}$ about the predicted outcome (AU), there is uncertainty about which model even is the correct one (EU).
Quantifying EU thus demands a bi-level framework that accounts for different hypotheses, 
uncertainty rising with the number of hypotheses deemed plausible \citep{hofman_2024_QuantifyingAleatorica}. 
The EU representation problem concerns this multiplicity and admits two competing viewpoints: Maximum uncertainty is attained when (V1) all hypotheses are equally likely \textit{or} (V2) only hypotheses reflecting full confidence for one label each are assigned nonzero probability (in complete disagreement).
Take the toy example of predicting a coin toss with EU concerning coin bias. 
Should EU be highest when the model deems any degree of bias equiprobable, or when it is sure the coin shows either heads or tails always but not which?
Clarity about this worst-case scenario is necessary in uncertainty-based tasks like active learning \citep{smith_prediction-oriented_2023} and out-of-distribution (OOD) detection \citep{azizmalayeri2024mitigatingoverconfidence}.
V1 fits a classical Bayesian view of associating uninformed beliefs with uniformity \citep{dubois_1996_RepresentingPartial}, 
while EU as per the entropy decomposition embodies V2, becoming maximal when all hypotheses express full confidence in utter disagreement \citep{shoja_2017_UncertaintyInformation, wimmer_2023_QuantifyingAleatoric}.

\color{black}
We do not aspire to resolve the philosophical dilemma but argue instead that the two EU manifestations might arise from different scenarios.
More precisely, we postulate that conflicting hypotheses (V2) emerge in the presence of \textit{shortcut learning (SCL)}:
Since there is only one true mapping from features to targets, one hypothesis at most---or even none---can have recovered this relationship, while the others necessarily point to spurious patterns.
Our experiments suggest that shortcuts in the data indeed prompt such disagreement.
This observation bears important insights for learning dynamics and has not, to the best of our knowledge, been studied despite ample discussion about robust generalization \citep{nagarajan2024understandingfailure, wald2022calibrationoutofdomain, richens2024robustagents}.
We emphasize that we do not claim SCL is the sole cause of conflicting hypotheses (see Sec.~\ref{sec:discussion}); rather, our results should be taken as encouraging further research in this direction.
\color{black}

\section{Background \& Related Work}
\label{sec:background}

\paragraph{Uncertainty Quantification}
\label{subseq:uq}

Bayesian methods come with a natural bi-level uncertainty representation, posing a second-order probability distribution $Q$ over first-order distributions $\bm{\theta}$ induced by hypotheses $h$ (for details, see App.~\ref{subsec:app-notation}),
and have therefore risen to gold standard in UQ \citep{izmailov_2021_WhatAre}.
We consider the special case of finite ensembles, which can be viewed as approximately Bayesian \citep{wilson_2020_BayesianDeep, wild_2023_RigorousLink, mlodozeniec_2024_ImplicitlyBayesian}, but our argumentation holds for any approach expressing EU via multiple predictions \citep[induced by a distribution or set of hypotheses;][]{hofman_2024_QuantifyingAleatorica} per instance.
Following the entropy ($H(\cdot)$) decomposition popularized by  \citet{houlsby_2011_BayesianActive} and \citet{kendall_2017_WhatUncertainties}, EU in the Bayesian setting is measured via \textit{mutual information}:
\begin{equation}
    \text{EU} = \underbrace{I(Y; \Theta)}_{\text{mutual information}} = \underbrace{H(\E_Q[Y|\Theta])}_{\text{entropy (TU)}} - \underbrace{\E_Q[H(Y|\Theta)]}_{\text{conditional entropy (AU)}} = \E_Q \left[D_{\text{KL}} \left(p(Y|\bm{\theta}) ~\|~ \bar{\bm{\theta}} \right) \right], 
    \label{eq:eu}
\end{equation}
where $\Theta$ denotes the random variable of first-order probability distributions.
Mutual information is equivalent to \textit{Jensen-Shannon divergence} in the finite-ensemble case.
The emphasis on disagreement is obvious from the Kullback-Leibler divergence term: $D_{\text{KL}}(\cdot)$ increases with deviation between individual hypotheses $p(Y|\bm{\theta})$ and the consensus prediction $\bar{\bm{\theta}}$ \citep[Eq.~\ref{eq:mi}, \ref{eq:consensus}-\ref{eq:consensus-ens};][]{shoja_2017_UncertaintyInformation}.

\paragraph{Shortcut Learning}
\label{subseq:scl}

SCL is fundamentally a problem of distribution shift and occurs when patterns picked up in training do not carry over to OOD scenarios 
\citep[see][for a comprehensive discussion]{steinmann_2024_NavigatingShortcuts}.
Relevant works discern \textit{stable} and \textit{unstable} features $X_s, X_u \subseteq X$, with $X_u$ non-causally correlated to $Y$, in varying terminology \citep[e.g.,][]{chalupka2015visualcausal, eastwood2023spuriositydidn}. 
Those distinctions imply the desirability of predicting $Y$ from $X_s$\footnote{
Unstable features, while poor predictors on their own, can still boost performance \citep{eastwood2023spuriositydidn}.
},
which is not always possible\footnote{
\citet{jalaldoust2024partialtransportability} argue that situations of \textit{non-transportability}, when $X_s$ is not available and more (of the same) data cannot alleviate the shortcut problem, amount to a form of irreducible AU.
}.
Relying on $X_u$ instead induces \textit{shortcuts} that work during training but break at deployment in OOD environments.
Some shortcuts
mirror real-world spurious correlations\footnote{
Associating cows with grass is reasonable but useless if failing to identify cows in other surroundings.},
others are introduced during data collection. 
The patterns can be subtle---e.g., high-frequency noise invisible to the human eye---and easily go undetected. 
Shortcuts abound across data modalities and affect downstream concerns like adversarial robustness and fairness \citep{geirhos2020shortcutlearning}.

\paragraph{Bridging the Gap}

Now, what makes models succumb to shortcuts?
\citet{steinmann_2024_NavigatingShortcuts} list ill-defined tasks and noisy $X_s$ as potential causes. 
More importantly, the same inductive biases we praise for enabling (in-distribution) generalization also encourage SCL. 
Neural networks (NNs) are especially susceptible: In extracting latent features that a
fully-connected module can digest, it is only rational under Occam's principle to rely on shortcuts when those induce the simplest risk-minimizing data representation\footnote{
Strong information compression from input to latent space can signal shortcuts \citep{adnan_2022_MonitoringShortcut}.
} \citep{geirhos2020shortcutlearning, friedrich_2023_TypologyExploringa}.
Some authors argue even that SCL cannot safely be avoided without a causal framework \citep{scholkopf_2021_CausalRepresentation}.
Considerable effort has been made in this spirit by moving beyond standard supervised settings.
Notably, 
a strand of recent work exploits the inherent multi-basin dynamics of ensembles with additional diversity-boosting components (e.g., special regularizers \citep{teney_evading_2022} or exposure to OOD training data \citep{pagliardini_agree_2023, scimeca_2024_MitigatingShortcut}).
\textcolor{red}{}
In that sense, we connect the dots between a principled discussion 
about disagreement to represent EU and algorithms using some pragmatic notion of uncertainty as a practical tool.

\section{Predictive Uncertainty in the Presence of Shortcut Learning}
\label{sec:results}



In short, we observe that uncertainty estimates based on data with explicit shortcuts differ from those without. 
We consider classification tasks derived from MNIST \citep[][pooling digits 1-3, 4-6, 7-9]{lecun_1998_GradientBasedLearning} and solved using deep ensembles of size 3 \citep[][
see App.~\ref{subsec:app-experiments} for details]{lakshminarayanan_2017_SimpleScalable}.
Shortcuts of strength $s$ are introduced in $s$\% of samples by coloring entire images (CMNIST3) or adding $1 \times 1$ colored pixels (PMNIST3) according to class\footnote{
Our code is available at \repourl.
} (Fig.~\ref{fig:example-data}).

\begin{figure}[h]
    \centering
    \includegraphics[width=\linewidth]{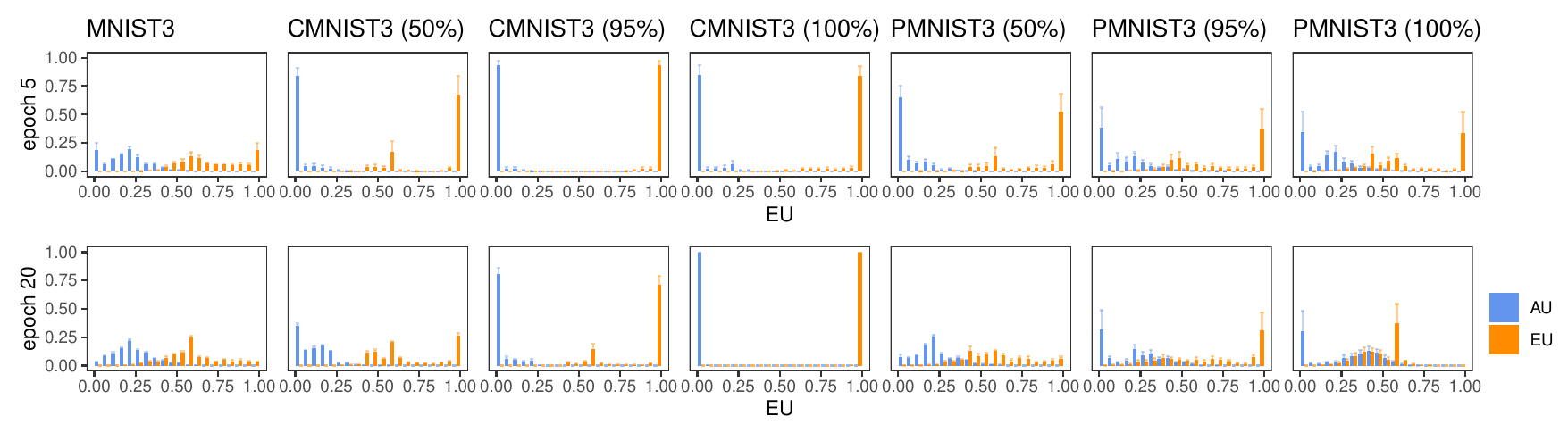}
    \caption{\textcolor{black}{Uncertainty estimates (20\% least-confident predictions) based on data with varying $s$; ensemble size 3; average over 3 independent runs (error bars: $\pm$ 1 standard error).}}
    \label{fig:eu-hist}
\end{figure}

Fig.~\ref{fig:eu-hist} shows the distribution of uncertainty estimates for MNIST3, CMNIST3 and PMNIST3 ($s \in \{50, 95, 100\}$).
The respective test data consist of original class-0 images not seen during training.
Uncertainty under SCL manifests as near-maximal mutual information\footnote{
We normalize all uncertainty components to values in [0, 1] according to Eq.~\ref{eq:entropy}.
} (orange bars) for CMNIST3, at near-full confidence (meaning zero AU; blue bars), under strong-enough shortcuts (columns 3-4).
In other words, the ensemble members produce conflicting predictions with high probability on different classes for most test images (less so when the shortcut is only partial).
This behavior follows the disagreement interpretation of EU (V2 in Sec.~\ref{sec:intro}).
There is a similar, if weaker, pattern for the less-pronounced {PMNIST3} cue.
Hypotheses start to agree again later in a sort of model collapse for PMNIST3 with $s = 100$ (note that estimates evolve as training progresses; see also App.~\ref{subsubsec:app-results}).
The shortcut-free MNIST3 task produces a diffuse distribution in both uncertainty components, reflecting varying degrees of confidence and disagreement across different test images (V1).
Only few samples provoke full disagreement.
Recall, however, that the test images are OOD for all datasets.
Using EU estimates as a signal in OOD detection would then yield quite different interpretations depending on whether SCL is at play: \textcolor{black}{trained on MNIST3, the evidence pointing toward OOD is rather weak, in contrast to training on data infused with a shortcut, even though both scenarios are, in fact, OOD.}
Taking uncertainty estimates at face value 
can thus be deceptive without a clear understanding of what EU via disagreement implies.

\color{black}
\section{Discussion}
\label{sec:discussion}

The evidence presented in Sec.~\ref{sec:results} suggests a relation between SCL and predictive uncertainty that has so far received little attention.
It does not, however, imply a one-to-one correspondence between SCL and high EU.
For one, situations other than SCL might prompt conflicting hypotheses: when the data are truly ambiguous or even contradictory, we must expect models to pick up different signals.
That said, such poor-quality data should manifest in higher AU and thus lower EU (due to the additivity of the components and the entropy being upper-bounded), reflecting the entanglement between both sources of uncertainty in finite-data settings pointed out by \citet{wimmer_2023_QuantifyingAleatoric}.  
On the other hand, SCL might occur without any conflict if all hypotheses come to agree on the same, shortcut-driven mapping.
Our results provide initial insights in what seems to be a promising direction, and underline the importance of studying the impact of distribution shifts on phenomena usually viewed in the i.i.d.\ vacuum. 
\color{black}

\section{Conclusion}

Much of recent research in UQ concerns methodological improvement. 
While indisputably important, this narrow focus---when it entails silent acceptance of assumptions like i.i.d.\ data---isolates the field from more fundamental discussions.
The confusion about the appropriate representation of EU arises from such a gap.
\textcolor{black}{We take a step toward reconciling} different views on representing EU (conflict \textit{vs} ignorance) by showing how shortcuts, which would not exist if the i.i.d.\ assumption actually held, affect uncertainty estimates in support of the conflict-type notion.
With this, we hope to contribute to a comprehensive view on generalization that stands the test of real-world learning situations.

\subsubsection*{Acknowledgments}
LW is supported by the DAAD program Konrad Zuse Schools of Excellence in Artificial Intelligence, sponsored by the German Federal Ministry of Education and Research.

\bibliography{references}
\bibliographystyle{abbrvnat}

\newpage
\appendix
\section{Appendix}

\subsection{Notation}
\label{subsec:app-notation}

In the following, we collect and define notational symbols used throughout the paper.

\subsubsection{General}
\label{subsubsec:notation-general}

\phantom{foo}

\renewcommand{\arraystretch}{1.5} 
\begin{tabular}{p{0.6\linewidth}  p{0.35\linewidth}}
    $\R$ & set of real numbers \\
    $\N$ & set of natural numbers \\
    $\mathcal{X} \subseteq \R^{d_x}, {d_x} \in \N$ & feature (or input) space \\
    $\mathcal{Y} \subseteq \R^{d_y}, {d_y} \in \N$ & target (or label, output) space \\ 
    $\mathcal{D} \in (\mathcal{X} \times \mathcal{Y})^n, n \in \N$ & set of training data \\
\end{tabular}

\subsubsection{Random Variables}
\label{subsubsec:notation-rv}

\phantom{foo}

\renewcommand{\arraystretch}{1.5} 
\begin{tabular}{p{0.6\linewidth}  p{0.35\linewidth}}
    $\E_{q}(\cdot)$ & expectation w.r.t. distribution $q$ \\
    $D_{\text{KL}}(q \| \cdot)$ & Kullback-Leibler divergence from distribution $q$\\
    $H(\cdot)$ & Shannon entropy \\
    $I(\cdot; \cdot)$ & mutual information
\end{tabular}

\paragraph{Shannon Entropy}
The \textit{Shannon entropy} \citep{shannon_1948_MathematicalTheory} of a discrete random variable (RV) $A$ with realizations in a sample space $\Omega$ is defined as:
\begin{equation}
    H(A) = - \sum_{\omega \in \Omega} \omega \log \omega  ~ \in[0, \log |\Omega|],
    \label{eq:entropy}
\end{equation}
where the logarithm is typically set to base 2 in accordance with an information-theoretic bit interpretation.
Entropy captures the potential information gain from observing the realization of RVs.
Consequently, it is minimal for RVs whose distribution is a Dirac measure, since the outcome is all but certain \textit{a priori}, and maximal for uniformly distributed RVs \citep{cover_2006_ElementsInformation}.
Normalization of $H(A)$ by $(\log |\Omega|)^{-1}$ maps entropy values to the unit interval.

\paragraph{Mutual Information}
Entropic measures give rise to the \textit{mutual information} between two RV:
\begin{align}
    I(A; B) &= H(A) - H(A | B) \notag \\
    &= D_{\text{KL}} \left(p(A, B) ~\|~ p(A) \otimes p(B) \right) \notag \\
    &= \E_{p(B)} \left[ D_{\text{KL}} \left(p(A|B) ~\|~ p(A)\right) \right],
    \label{eq:mi}
\end{align}
with $\otimes$ denoting the outer product distribution.
Mutual information is a measure of statistical independence and quantifies how much information can be gained about $A$ by observing $B$, or \textit{vice versa}.
It vanishes at perfect independence, i.e., when the joint distribution factorizes into $p_A \otimes p_B$, and realizations from $A$ do not decrease uncertainty over $B$.
Alternatively, we can view $I(A; B)$ as the expected divergence between the conditional $p(A|B)$ and the marginal $p(A)$ that increases with more information in $B$ about $A$
\citep{cover_2006_ElementsInformation}.
This latter interpretation is particularly useful to understand the emphasis on disagreement between base learner and consensus prediction expressed in Eq.~\ref{eq:eu} and Eq.~\ref{eq:eu_full}.

\newpage
\subsubsection{Uncertainty Decomposition}
\label{subsubsec:notation-uq}

We largely adopt the notation of \citet{hofman_2024_QuantifyingAleatorica} in the following.

\begin{tabular}{p{0.6\linewidth}  p{0.35\linewidth}}
    $\triangle_K, K \in \N$ & $K-1$ simplex \\
    $\mathbb{P}(\mathcal{Y})$ & set of first-order probability distributions over $\mathcal{Y}$ \\
    $\mathcal{H} = \{h: \mathcal{X} \rightarrow \mathbb{P}(\mathcal{Y}) ~|~ h \text{ is of a certain functional form}\}$ & space of probabilistic hypotheses \\
    $Q: \mathcal{H} \rightarrow [0, 1]$ & second-order probability distribution \\
    $Y$ & RV of outcome labels \\
    $\Theta$ & RV of first-order distributions \\
    $M \in \N$ & ensemble size
\end{tabular}

\paragraph{Bi-Level Uncertainty Representation}
For the classification case discussed here, we assume first-order label distributions $\bm{\theta} = p(Y | \cdot)$, , where $Y$ denotes the random outcome variable, to be categorical with $K = |\mathcal{Y}| < \infty$ possible outcomes, equating the set $\mathbb{P}(\mathcal{Y})$ of such distributions with the $K-1$ simplex $\triangle_K = \{\bm{\theta} \in [0, 1]^K: \| \bm{\theta} \|_1 = 1\}$.

Bayesian agents produce a probability distribution over the probabilistic prediction $\bm{\theta}$ (\textit{posterior predictive density; PPD}) that is induced by the second-order distribution $Q$.
$Q$ assigns probabilities to hypotheses from $\mathcal{H}$.
In the Bayesian paradigm, a prior belief $Q(\cdot)$ is updated to a posterior belief $Q(\cdot | \mathcal{D})$ after observing data, giving rise to the following PPD:
\begin{equation}
    p(\bm{\theta}) = \int_{\mathcal{H}} \1_{h{(\bm{x}) = \bm{\theta}}} \text{ d} Q(h| \mathcal{D}).
\end{equation}
Here, $h(\bm{x}) \in \triangle_K$ models the ground-truth conditional density $\bm{\theta}^\ast = (\theta_1^\ast, \dots, \theta_K^\ast)^\top$ with $\theta_k^\ast =$ $p(Y = k | \bm{x})$.
We will sometimes omit the conditioning on $\bm{x}$ so as to not overload notation.

\paragraph{Bayesian Model Average}
The Bayesian paradigm further admits a first-order predictive distribution as an expectation over all possible models (hypotheses), yielding the \textit{consensus prediction}
\begin{equation}
    \bar{\bm{\theta}} = \int_{\mathcal{H}} h(\bm{x}) \text{ d} Q(h| \mathcal{D}).
    \label{eq:consensus}
\end{equation}
In most practical problems, both $Q$ and the integral in Eq.~\ref{eq:consensus} are intractable.
This issue is typically addressed by (unbiased) Monte Carlo integration over samples from $Q$ \citep[as obtained by some approximately Bayesian---e.g., sampling-based or variational---method;][]{andrieu_2003_IntroductionMCMC}.
We specifically consider ensembles with $M$ base learners \citep{wilson_2020_BayesianDeep}, leading to the following approximation:
\begin{equation}
    \bar{\bm{\theta}} \approx \frac{1}{M} \sum_{m = 1}^M h^{[m]}(\bm{x}).
    \label{eq:consensus-ens}
\end{equation}
Note that Eq.~\ref{eq:consensus-ens} is only a valid approximation of Eq.~\ref{eq:consensus} if all ensemble members represent the same structural form of hypothesis \citep{minka_2002_BayesianModel}.
This is the case for deep ensembles \citep{lakshminarayanan_2017_SimpleScalable}, where base learners differ solely by parameterization as a consequence of random weight initialization and stochastic training elements. 

\paragraph{Entropy Decomposition}

With $\Theta$ denoting the RV whose realizations are distributions $\bm{\theta} \in \triangle_K$, we can derive the components of predictive uncertainty as
\begin{equation}
    \underbrace{H(Y)}_{\text{TU}} = \underbrace{H(Y | \Theta)}_{\text{AU}} + \underbrace{I(Y; \Theta)}_{\text{EU}}.
    \label{eq:decomp}
\end{equation}

\begin{itemize}
    \item[(TU)] The \textit{total} uncertainty of a prediction obtained via Bayesian model averaging (Eq.~\ref{eq:consensus}) is quantified via Shannon entropy (Eq.~\ref{eq:entropy}) and defined as
    \begin{equation}
        H(Y) = H \left(\E_Q \left[Y | \Theta \right] \right) = H(\bar{\bm{\theta}}) = - \sum_{k = 1}^K \bar{\theta}_k \log \bar{\theta}_k.
        \label{eq:tu_full}
    \end{equation}
    The more $\bar{\bm{\theta}}$ concentrates on a single outcome (pushing it toward one of the simplex corners), the lower its corresponding uncertainty. 
    \item[(AU)] Similarly, we obtain \textit{aleatoric} uncertainty as the \textit{conditional entropy} of the outcome:
    \begin{equation}
        H(Y | \Theta) = \E_Q \left[H(Y | \Theta) \right] = -\int p(\bm{\theta}) H(\bm{\theta}) \text{ d} \bm{\theta}.
        \label{eq:au_full}
    \end{equation}
    \item[(EU)] The \textit{epistemic} component emerges as the residual quantity from the additive decomposition of Eq.~\ref{eq:tu_full}, which amounts to the mutual information between $Y$ and $\Theta$:
    \begin{align}
        I(Y; \Theta) &= H(Y) - H(Y | \Theta) \notag \\
        &= \E_Q \left[D_{\text{KL}}(p(Y|\bm{\theta}) ~\|~ p(Y) \right] \notag \\
        &= \E_Q \left[D_{\text{KL}} \left(p(Y|\bm{\theta}) ~\|~ \bar{\bm{\theta}} \right) \right].
        \label{eq:eu_full}
    \end{align}
\end{itemize}

\subsection{Experiments}
\label{subsec:app-experiments}

\subsubsection{Experimental Details}
\label{subsubsec:app-experiments}

\paragraph{Datasets}

We consider tasks based on MNIST \citep{lecun_1998_GradientBasedLearning}. 
Training sets comprise 10k images with balanced classes:
\begin{itemize}
    \item \textbf{MNIST3.} 3-class MNIST version where original classes are pooled into classes $\{0, 1, 2\}$, consisting of digits 1-3, 4-6, 7-9, respectively (leaving class 0 as OOD test data).
    \item \textbf{CMNIST3.} Version of MNIST3 with global shortcut coloring digits 1-3 in red, digits 4-6 in green, and digits 7-9 in blue  
    \citep[similar experiments are conducted in, e.g.,][]{jalaldoust2024partialtransportability}.
    \item \textbf{PMNIST3.} Version of MNIST3 with local shortcut adding a colored $1 \times 1$ patch in the top left of each image; digits 1-3: red, digits 4-6: green, digits 7-9: blue \citep[similar experiments are conducted in, e.g.,][]{adnan_2022_MonitoringShortcut}.
    \item \textbf{MNIST0.} Class 0 from original MNIST data, used as OOD data.
\end{itemize}
We vary the shortcut strength 
by modifying $s$\% of images (e.g., CMNIST3 with $s = 50$ is created by coloring 50\% of MNIST3 images and leaving the rest black-and-white).

\paragraph{Models}

We use deep ensembles \citep{lakshminarayanan_2017_SimpleScalable} of small NNs with one convolutional layer (16 filters of size $3 \times 3$), followed by max-pooling and two fully-connected layers (dimensions $16 \cdot 196$ and $128$, respectively), ReLU activations in the hidden layers, and softmax activation in the final layer.

\paragraph{Training}

We train our models for 25 epochs with AdamW optimization \citep{loshchilov_2019_DecoupledWeight}, an initial learning rate of $0.001$ that is reduced by a factor of $0.1$ if it plateaus for 10 consecutive epochs, and weight decay of $0.01$.
Batch size is set to $128$.

\paragraph{Software}

Our code is mainly based on \texttt{PyTorch} \citep{paszke_2019_PyTorchImperative} and \texttt{PyTorch Lightning} \citep{falcon_2023_PyTorchLightning}.
We performed all experiments on CPU and make our code available at \repourl.

\subsubsection{Further Results}
\label{subsubsec:app-results}

For all results, we normalize the uncertainty components to values in $[0, 1]$ (using the upper bound on Shannon entropy, Eq.~\ref{eq:entropy}, given by the number of possible outcomes).

\paragraph{More Epochs}

\begin{figure}[ht]
    \centering
    \includegraphics[width=\textwidth]{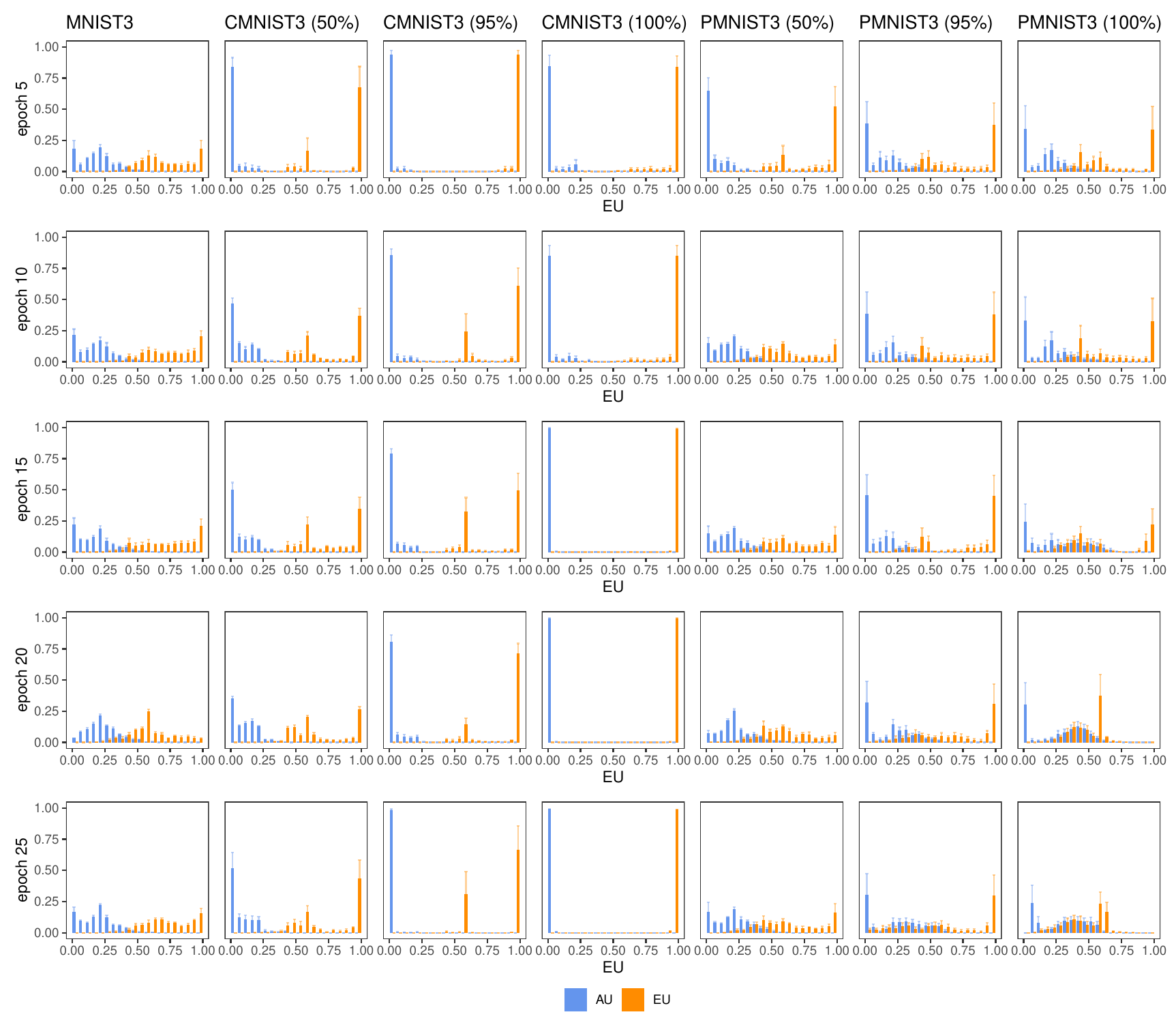}
    \caption{\textcolor{black}{\textit{More epochs.} Uncertainty estimates (20\% least-confident predictions) based on data with varying $s$; ensemble size 3; average over 3 independent runs (error bars: $\pm$ 1 standard error).}}
    \label{fig:alleps}
\end{figure}

Fig.~\ref{fig:alleps} shows results on MNIST0 test data for some more training epochs in addition to Fig.~\ref{fig:eu-hist}.
We observe that model weights and their magnitude changing over the course of training also affect the depicted uncertainty estimates.

\paragraph{Effect of Shortcut Strength}

For CMNIST3, some effect is visible even when only half of the training images are colored (column 2 in Fig.~\ref{fig:alleps}).
PMNIST3 (column 5) produces uncertainty estimates as diffuse as for the MNIST3 (column 1) data without any shortcut; disagreement-seeking behavior only occurs for strong shortcuts.
In the case of a perfect spurious correlation, the CMNIST3-trained learner (column 4) settles for full confidence and full disagreement on virtually all test samples.
PMNIST3 (column 7) exhibits a sort of model collapse, where hypotheses start to agree more in later epochs.
Stronger correlations also lead to lower accuracy\footnote{
Accuracy is calculated with MNIST3 as a test set (i.e., images from the same classes as in the training data, but free of shortcuts---accuracy on the never-seen OOD class would be consistently 0).
} (Tab.~\ref{tab:perf_weakstrong}) as the models come to rely on the shortcut and pick up little of the stable pattern.
For instance, the learner trained on CMNIST3 achieves 95\% accuracy (epoch 25) when exposed to 50\% shortcut strength, nearly on par with MNIST3, but makes heavy use of the 100\% coloring cue and deteriorates to 41\% accuracy in the strong-shortcut scenario.

\begin{table}[ht]
    \scriptsize \color{darkgray}
    \centering
    \begin{tabular}{|r|r|r|r|r|r|r|r|}
        \hline
        Epoch & MNIST3 & CMNIST3 50 & CMNIST3 95 & CMNIST3 100 & PMNIST3 50 & PMNIST3 95 & PMNIST3 100\\
        \hline
        5 & 0.84 (0.05) & 0.86 (0.01) & 0.62 (0.02) & 0.44 (0.04) & 0.72 (0.11) & 0.68 (0.11) & 0.57 (0.06)\\
        \hline
        10 & 0.95 (0.00) & 0.90 (0.00) & 0.61 (0.01) & 0.44 (0.04) & 0.86 (0.06) & 0.67 (0.11) & 0.57 (0.06)\\
        \hline
        15 & 0.96 (0.00) & 0.93 (0.00) & 0.72 (0.01) & 0.41 (0.02) & 0.87 (0.06) & 0.67 (0.11) & 0.56 (0.06)\\
        \hline
        20 & 0.97 (0.00) & 0.95 (0.00) & 0.67 (0.01) & 0.41 (0.02) & 0.86 (0.06) & 0.67 (0.11) & 0.68 (0.11)\\
        \hline
        25 & 0.97 (0.00) & 0.95 (0.00) & 0.67 (0.03) & 0.41 (0.03) & 0.87 (0.06) & 0.67 (0.11) & 0.68 (0.11)\\
        \hline
    \end{tabular}
    \caption{Accuracy and corresponding standard errors over 3 independent runs; ensemble size 3.}
    \label{tab:perf_weakstrong}
\end{table}

\paragraph{Larger Ensemble Size}

\begin{figure}[ht]
    \centering
    \includegraphics[width=\textwidth]{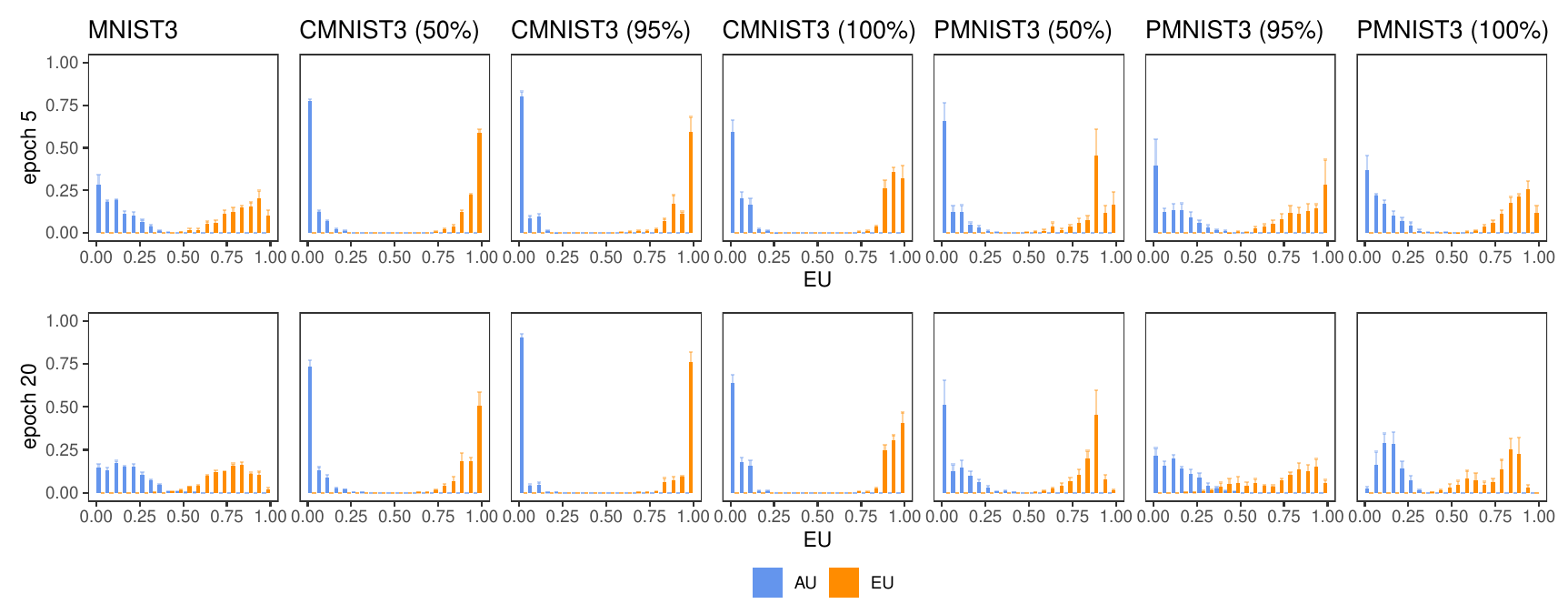}
    \caption{\textcolor{black}{\textit{Larger ensemble.} Uncertainty estimates (20\% least-confident predictions) based on data with varying $s$; ensemble size 5; average over 3 independent runs (error bars: $\pm$ 1 standard error).}}
    \label{fig:enssize}
\end{figure}

When we increase the ensemble size to 5 (Fig.~\ref{fig:enssize}, as opposed to size 3 for the results reported in Fig.~\ref{fig:eu-hist}, with otherwise identical settings), we observe a similar tendency for CMNIST3 toward strong disagreement.
The PMNIST3 training still provokes more conflict than shortcut-free MNIST3 but to a lesser degree than CMNIST3.
In general, with growing ensemble size, the probability of some members converging to the same predictions rises.
Note that it is no longer possible for all hypotheses to settle on completely conflicting predictions: Since the number of ensemble members is now larger than (and not a multiple of) the number of classes, the maximum EU value of 1 is not attainable in this setting\footnote{
EU values may still end up in the highest bin of (0.95, 1]: 5 predictions for some class at full confidence each, with at most two members agreeing at a time will produce some perturbation of a (0.2, 0.4, 0.4) class probability distribution.
This distribution has 0.96 entropy (TU), while AU, as the average entropy over one-hot probability vectors, is 0.
Per the additivity constraint of the entropy decomposition, this leaves EU = TU - AU at 0.96, and thus in the last bin.
}.

\color{black}
\paragraph{In-Distribution Results}

\begin{figure}[ht]
    \centering
    \includegraphics[width=\textwidth]{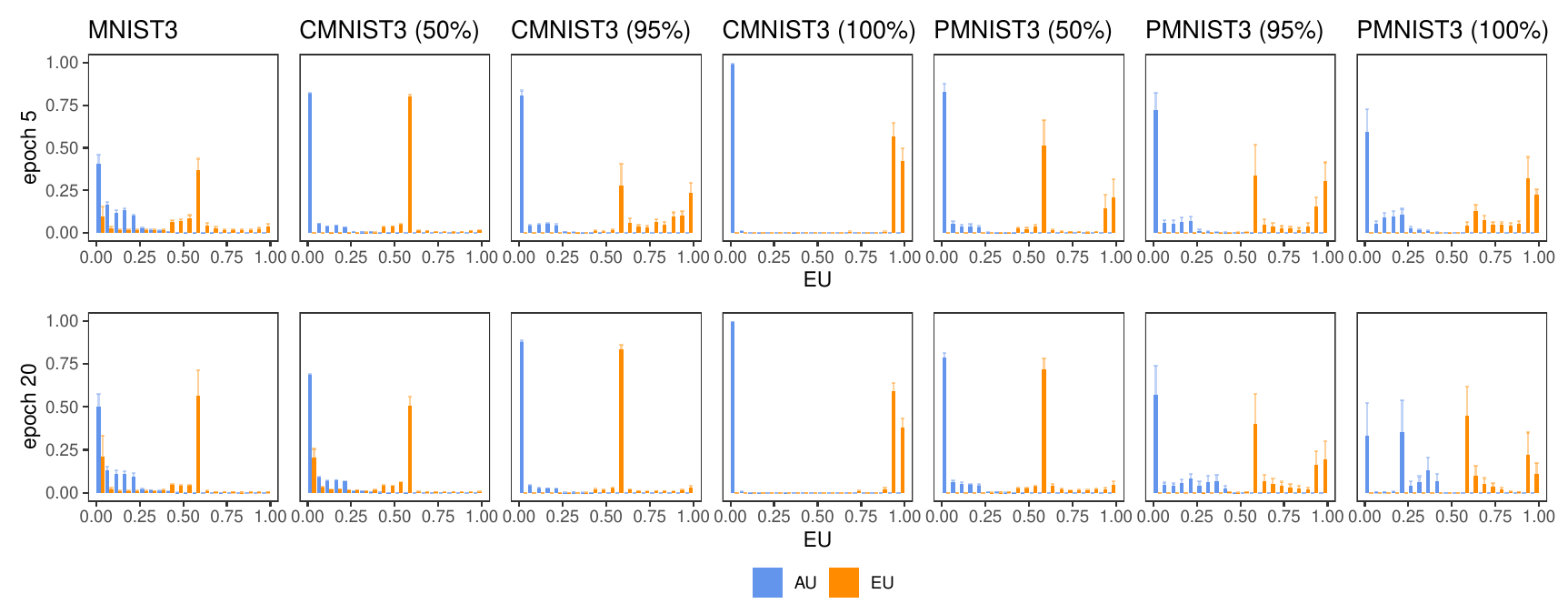}
    \caption{\textcolor{black}{\textit{In-distribution test data.} Uncertainty estimates (20\% least-confident predictions) based on data with varying $s$, where train and test classes coincide but shortcuts are only present in the training data; ensemble size 3; average over 3 independent runs (error bars: $\pm$ 1 standard error).}}
    \label{fig:ind}
\end{figure}

Finally, we provide in Fig.~\ref{fig:ind} the results corresponding to Fig.~\ref{fig:eu-hist} with in-distribution test data (in the sense that both partitions contain the same classes, as opposed to the OOD experiments based on predictions for the held-out 0 class). 
Shortcuts are present only during training.
As might be expected, the disagreement between hypotheses is generally less pronounced, suggesting that the models pick up some of the non-spurious patterns.
For CMNIST3 with perfect shortcut, we still see high EU values (note, though, that Fig.~\ref{fig:ind} only depicts results for the 20\% most-uncertain predictions).

\end{document}